\documentclass[11pt]{article}
\usepackage[preprint]{acl}
\usepackage{times,latexsym}
\usepackage[T1]{fontenc}
\usepackage[utf8]{inputenc}
\usepackage{microtype,inconsolata,graphicx,colortbl,booktabs,tabularx,array,amsmath,placeins,needspace}
\renewcommand{\footnoterule}{\kern-3pt \hrule width 2in \kern 2.6pt}
\newcolumntype{Y}{>{\raggedright\arraybackslash}X}
\newcommand{\AbstractResults}{Across five rankings of the three primary rules, the signed mean held-out-accuracy difference (InternVL 2B minus Self-check) was +6.6\% on shape (descriptive range +2.2\% to +12.3\%) and +1.5\% on color (range -2.1\% to +8.1\%). The corresponding Random minus Self-check differences were +2.3\% on shape and -0.6\% on color. These ranges summarize the five ranking-seed results.}
\newcommand{\DevelopmentResults}{On the 288 update-selection candidates, Qwen 2B falsely admits 89.9\% of incorrect shape answers and 100.0\% of incorrect color answers; the corresponding InternVL 2B rates are 31.9\% and 69.1\%. Table~\ref{tab:development} reports the separate checker-measurement split as well.}
\newcommand{\SensitivityResults}{Removing images changes 20.5\% of Qwen 2B verdicts for shape and 6.2\% for color. The corresponding InternVL 2B rates are 61.5\% and 72.2\%. Shuffling changes 50.3\% and 57.3\% of InternVL 2B verdicts.}
\newcommand{\DownstreamResults}{Across five rankings, the signed mean held-out-accuracy difference from no update for InternVL 2B was +8.5\% on shape (range +6.5\% to +11.0\%; positive in 5 of 5 sets) and +1.0\% on color (range -0.7\% to +3.6\%; positive in 4 of 5 sets). The original-seed controls are reported in Table~\ref{tab:downstream}; these checkpoint-specific comparisons do not estimate a size effect.}
\newcommand{\DirectComparatorResults}{Across five rankings, the signed mean held-out-accuracy difference (Random minus Self-check) was +2.3\% on shape (range -1.6\% to +9.0\%; positive in 3 of 5 sets) and -0.6\% on color (range -3.5\% to +3.4\%; positive in 2 of 5 sets). The corresponding InternVL 2B minus Self-check differences were +6.6\% on shape (range +2.2\% to +12.3\%; positive in 5 of 5 sets) and +1.5\% on color (range -2.1\% to +8.1\%; positive in 3 of 5 sets). These descriptive summaries condition on the realized selected sets and fitted seeds.}
\newcommand{\SelectionResults}{Primary selection rules use five ranking seeds. Across those sets, InternVL 2B retains 37--45 correct proposer targets on shape and 37--41 on color; self-check ranges are 33--40 and 35--42. Table~\ref{tab:selection} reports the full primary and control ranges. Because the rules select different examples, these summaries do not identify the cause of held-out accuracy differences.}
\newcommand{\ReliabilityUtilityResults}{The five primary ranking sets create repeated complete-rule points with different selected-target precision and different downstream outcomes. For example, InternVL 2B's shape precision ranges from 77.1\% to 93.8\%, while its signed mean held-out-accuracy difference from no update ranges from +6.5\% to +11.0\%. These paired observations do not identify target correctness as the mechanism.}

\newcommand{\ConclusionResults}{Across five rankings of the three primary gates, selected-target precision and held-out accuracy did not yield a consistent ordering; the single-seed controls broaden the rule comparison without adding ranking replication. On the same selected IDs, the signed mean held-out-accuracy difference (program target minus proposer target) for Self-check was +18.4\% on shape and +8.6\% on color in the original-seed diagnostic. The results support reporting gate correctness, selected-set composition, and downstream utility separately.}

\title{Measuring Answer-Gate Selection in Multimodal Self-Training:\\Held-Out Accuracy on Generated Counting Tasks}
\author{
  Mengzhe Geng\thanks{\raggedright Corresponding author: \href{mailto:Mengzhe.Geng@nrc-cnrc.gc.ca}{\textcolor{black}{\textnormal{Mengzhe.Geng@nrc-cnrc.gc.ca}}}}\\
  National Research Council Canada
  \And
  Guanqi Zhu\\
  Independent Researcher
  \AND
  Yehui Tang\\
  Imperial College London
  \And
  Xin Chen\\
  The Chinese University of Hong Kong
}
\begin{document}
\hypersetup{linkcolor=black}
\maketitle
\hypersetup{linkcolor=darkblue}
\begin{abstract}
Model-generated answers can expand multimodal training data but can also turn wrong answers into supervision. A gate is a complete rule for deciding which candidates enter an update; a checker is one model-based component of that rule. A gate can reject wrong answers while changing which examples enter training, so gate quality and update utility are separate measurements. We study this measurement problem by comparing six binary admission rules under a fixed proposer, learner, update recipe, one-epoch schedule, and 48-example budget on two program-generated counting tasks. Four fixed checkpoints score candidates; the three primary rules use five ranking seeds and three fitted seeds per set, while controls use the original ranking seed. \AbstractResults{} In an original-seed diagnostic with three fitted updates per target, replacing proposer targets with program targets yielded signed mean held-out-accuracy differences of +18.4\% on shape and +8.6\% on color for Self-check. The results show why answer admission, selected-target precision, and downstream update utility should be measured separately.
\end{abstract}

\section{Introduction}
Self-training turns a model's generated answers into labels.\footnote{\raggedright Public reproducibility repository:\\{\hypersetup{urlcolor=black}\url{https://github.com/MENGZHEGENG/vision-agent-interventional-audit}}.} A checker can reject incorrect labels, but its selection rule also changes which examples enter training. The measurement question is whether a gate that admits more correct answers also produces a more useful update. We measure candidate-level agreement with program answers, selected-target precision, and held-out accuracy separately.

We fix the candidate pool, proposer, learner, and update recipe, then compare complete rules across repeated rankings. Correctness of admitted answers and utility of their updates remain distinct estimands because selection changes the consumed population.

\begin{figure*}[t]
\centering
\includegraphics[width=\textwidth]{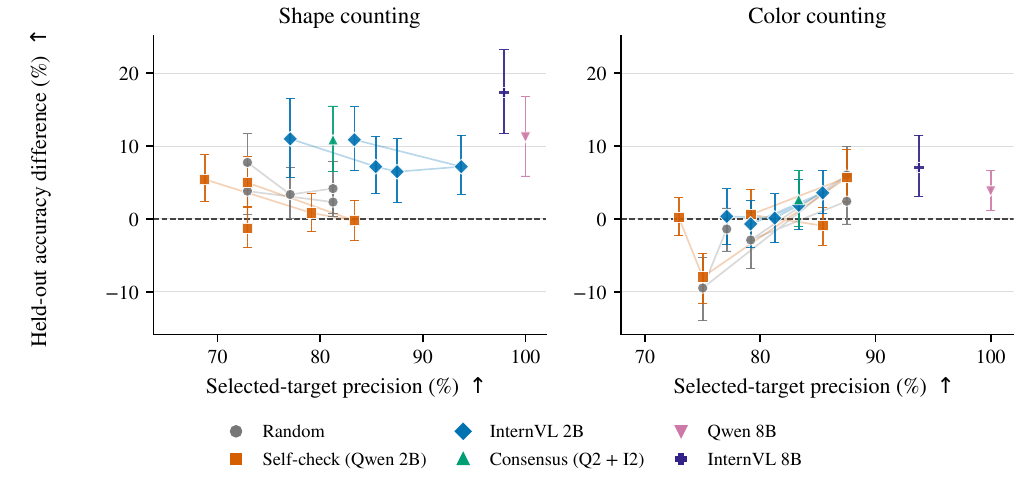}
\caption{Selected-target precision and signed held-out accuracy difference from no update (update $-$ no update, \%). The three primary rules contribute five points each, one per ranking seed; original-seed control arms contribute one point at the original seed. Error bars are conditional 95\% scene-group intervals for each realized set. Colors and markers identify the same rules in both tasks. Q2 is Qwen3-VL-2B-Instruct, and I2 is InternVL3.5-2B-Instruct.}
\label{fig:relationship}
\end{figure*}

Figure~\ref{fig:relationship} pairs consumed-target correctness with held-out accuracy for complete rules. It does not identify a causal effect of target correctness.
\paragraph{Contributions.} We (1) compare four binary answer checkers on a shared generated-candidate pool, (2) evaluate complete admission rules under equal 48-example updates, and (3) report finite-seed relationships among gate measurements, consumed-set composition, and held-out utility, alongside a frozen-set target-source diagnostic. We hold the proposer, learner, and update recipe fixed to measure how complete gate rules change admission and update outcomes.

\section{Related Work}
\paragraph{Multimodal judging.} Studies measure compositional and informativeness biases in multimodal judgments \citep{mmjudgebias2026,birch2026}. Visual perturbation and filtering also diagnose hallucination and region evidence \citep{perceptualhallucination2026,csgr2026,cvpd2026}. VISE uses geometric and semantic invariance rewards to strengthen visual conditioning during self-evolution \citep{vise2026}. Our study measures answer-admission decisions and downstream updates; visual sensitivity alone does not assess answer correctness or update utility.

\Needspace{16\baselineskip}
\paragraph{Selecting visual supervision.} Self-Filter uses a vision-language model to select instruction data by difficulty and diversity, while DOSE scores multimodal examples with off-the-shelf models; both report downstream tuning results after changing the selected corpus \citep{selffilter2024,dose2026}. COINCIDE selects examples by concept--skill transferability and evaluates cross-task transfer, so its signal targets data value instead of admission of a proposed answer \citep{coincide2024}. Perception-Verified Self-Training verifies captions and answers and regenerates reasoning targets. Verifiable QA Generation for Self-Evolving Models (VQS) constructs program-verified structured records, derives questions and answers from them, and iteratively trains parser and solver models across ten benchmarks, with human evaluation \citep{perceptionverified2026,vqs2026}. Our study holds a generated candidate pool and learner recipe fixed, then measures complete admission rules on answer correctness, consumed-target precision, and held-out utility. BIRCH studies comparative judging and corrects image-inconsistent candidate answers with a truthful anchor, exposing informativeness bias \citep{birch2026}. STIC, counterfactual visual distillation, and multimodal reward-hacking studies examine preference construction, visual evidence, or supervision failures \citep{stic2024,cvpd2026,rewardhacking2026}. Other work studies self-correction, corrupted multimodal data, self-evolving rewards, or process-supervised distillation \citep{scl2025,corrupted2025,evolmm2026,selfjudge2026,ireasoner2026,trinity2026,rthreev2025,kopsd2026}.

\section{Protocol and Measurements}
\subsection{Candidate answers and admission}
For image--question input $x_i$, proposer $P$ returns integer answer $\hat y_i$. Binary checker $V_m$ receives the image, question, and that answer, and returns YES or NO. Proposer parsing requires an exact stripped integer from 0 to 4; checker parsing requires an exact YES/NO verdict. Malformed answers or verdicts are not admitted.

Program answer $y_i$ supports three separate measurements. Candidate-level admission precision is the fraction of accepted candidate answers that are correct. False admission is the fraction of parseable incorrect proposer answers that a rule accepts. Malformed proposer answers are never admitted and are excluded from this denominator; every candidate in both development splits was parseable in both tasks. Coverage is the fraction of all candidates accepted. For a fixed 48-example update, selected-target precision (proposer-answer precision among consumed examples) is the analogous fraction among the examples actually consumed. Their denominators differ. These precision measures describe supervision correctness at the observed coverage; they do not measure the downstream benefit of an update.

\subsection{Images and information access}
Checkers see original images, no image, or images shuffled within development split, query target, edit type, and question. Shuffled images come from different scene groups. This controls the textual query while changing its visual context. We measure verdict changes, without interpreting them as proof of visual reasoning.

The selector implementation omitted program answers, derived correctness fields, raw candidate responses, and raw verdict responses; program answers were joined only after selected IDs were frozen. However, reference answers remained in shared development records accessible to the broader pipeline, so we cannot rule out their availability outside the selector code.

The gate-comparison updates use only the image, question, and proposer answer from each selected example; program answers never serve as training targets in those arms. A separate frozen-set diagnostic replaces the proposer target with the program answer after selection, and is identified as such below. We score held-out answers against program answers only after inference.

\subsection{Equal-budget selection}
Each task uses a deterministic seeded random-ranking baseline implemented as the SHA-256 hash of \texttt{selection\_seed:sample\_id} and retains the first 48 distinct scene groups; the declared ranking input contains no program answers or derived correctness fields. The three primary rules (random, self-check, and InternVL 2B) are evaluated with five ranking seeds: 20261004, 20261013, 20261014, 20261015, and 20261016. Consensus and the two 8B gates are evaluated at the original seed 20261004 as original-seed controls. Consensus requires both primary checkers to accept. The ranking is shared across rules within a seed; eligibility and the resulting example composition still differ between rules.

\subsection{Downstream estimand}
Let binary $z_{rqasj}$ indicate a correct answer from rule $r$, ranking seed $q$, training seed $a$, scene group $s$, and variant $j$. Each task has three fitted seeds, 96 held-out groups, and three variants. For the no-update reference, $z_{bqasj}=z_{bsj}$ for every ranking seed $q$ and fitted seed $a$; the reference may vary across scene groups $s$ and variants $j$. For a realized ranking seed $q$, the accuracy difference from reference rule $b$ is
\begin{equation}
\Delta_{r,b}^{(q)}=\frac{1}{96}\sum_{s=1}^{96}\left[\frac{1}{9}\sum_{a=1}^{3}\sum_{j=1}^{3}(z_{rqasj}-z_{bqasj})\right].
\end{equation}
We report $100\Delta_{r,b}^{(q)}$ as a signed accuracy difference (new $-$ reference, \%). For each realized selection set, we report mean accuracy, the three fitted-seed accuracies, and a two-sided 95\% percentile interval from 20,000 resamples of the 96 group means. These intervals condition on that selected set and the three fitted seeds. Across the five primary ranking seeds, we report the unweighted mean, minimum and maximum, and the number of positive contrasts; these summaries do not estimate a population of selection seeds. Comparisons are descriptive and are not adjusted for multiplicity. An interval spanning zero does not establish equivalence.

\section{Experimental Setup}
\subsection{Generated tasks}
Shape counting and color counting use $256\!\times\!192$ scenes with six objects, three shape categories, and six colors. English counting templates ask about shapes or colors and have integer answers from 0 to 4. The scenes contain no people or demographic attributes; they do not assess other languages or population representativeness. Each scene group supplies clean, answer-changing, and answer-preserving variants. An answer-changing edit alters one object's queried attribute to increase or decrease the target count by one. An answer-preserving edit changes an unqueried attribute of a non-target object: its color in shape counting or its shape in color counting. Development contains 96 groups per task in two splits: the update-selection split and the checker-measurement split. Each split contains 288 inputs; updates use only the update-selection split, for 576 development inputs per task.

Development generation seeds are 20261003 for shape and 20261004 for color. Fresh held-out generation uses 20261011 and 20261012, respectively, with 96 groups and 288 examples per task. Distinct generation seeds and layout grids separate development and held-out scenes. Programmatic visual scenes follow the diagnostic visual question answering (VQA) tradition exemplified by CLEVR \citep{johnson2017clevr}; these tasks support controlled comparisons, not estimates of natural-image question answering.

\subsection{Checkpoints and rules}
The proposer and adapted learner are Qwen3-VL-2B-Instruct \citep{qwen3vl2025}. The four fixed checkers are Qwen3-VL-2B-Instruct, InternVL3.5-2B-Instruct, Qwen3-VL-8B-Instruct, and InternVL3.5-8B \citep{qwen3vl2025,internvl2025}. For concise labels, Q2, I2, Q8, and I8 refer to these checkpoints in that order. The first checker is the proposer itself; the InternVL checkers assess Qwen-generated answers, so they are cross-model gates. The size labels 2B and 8B denote nominal parameter counts in billions; they do not equate architecture, pretraining, or error dependence.\begin{table}[!b]
\centering
\caption{Update rules. All trainable rules consume 48 examples.}
\label{tab:rules}
{\nolinenumbers\small
\begin{tabularx}{\columnwidth}{@{}lY@{}}
\toprule\textbf{Rule}&\textbf{Eligibility condition}\\\midrule
No update&Unchanged starting weights\\
Random&Any valid proposer answer\\
Self-check&Q2 accepts\\
InternVL 2B&I2 accepts\\
Consensus&Q2 and I2 accept\\
Qwen 8B&Q8 accepts\\
InternVL 8B&I8 accepts\\\bottomrule
\end{tabularx}}
\end{table}

Table~\ref{tab:rules} defines all seven arms. The three primary rules are Random, Self-check (Q2), and InternVL 2B (I2); Consensus combines Q2 and I2 as an original-seed control. The two 8B gates are additional checkpoint-specific controls, not a controlled scaling experiment.

\subsection{Training and decoding}
Each trained update and seed starts from fresh base weights. Low-rank adaptation (LoRA)~\citep{hu2022lora} uses rank 8, alpha 16, dropout 0.05, and query, key, value, and output projections. AdamW uses learning rate $10^{-4}$, $\beta=(0.9,0.999)$, $\epsilon=10^{-8}$, and weight decay 0.01; gradient norm clipping is 1. Each update performs 48 singleton optimizer steps, consuming all selected examples exactly once. \par\Needspace{3\baselineskip} Each seed shuffles the training-example order. Adapters are merged before held-out evaluation.

\Needspace{13\baselineskip}
Shape training seeds are 20261005, 20261007, and 20261009; color seeds are 20261006, 20261008, and 20261010. Each primary ranking seed therefore has three fitted updates per rule; the controls use three fitted updates at the original ranking seed. The no-update condition reuses one deterministic prediction vector per task and serves as a shared reference, not as a fitted seed. Development uses deterministic decoding with at most 16 new tokens and exact parsing. Held-out evaluation allows 64 new tokens and parses the first standalone nonnegative integer; values outside the target range 0--4 are scored as incorrect. This held-out parser is more permissive than the candidate parser. In the five-seed comparison, 18 of 31,680 held-out predictions are outside 0--4 (all 5).

InternVL receives one resized $448\!\times\!448$ image with ImageNet normalization, without dynamic tiling.

The reported optimizer and LoRA settings were fixed; no hyperparameter search was run and no in-context examples were used.

\begin{figure*}[t]
\centering\includegraphics[width=\textwidth]{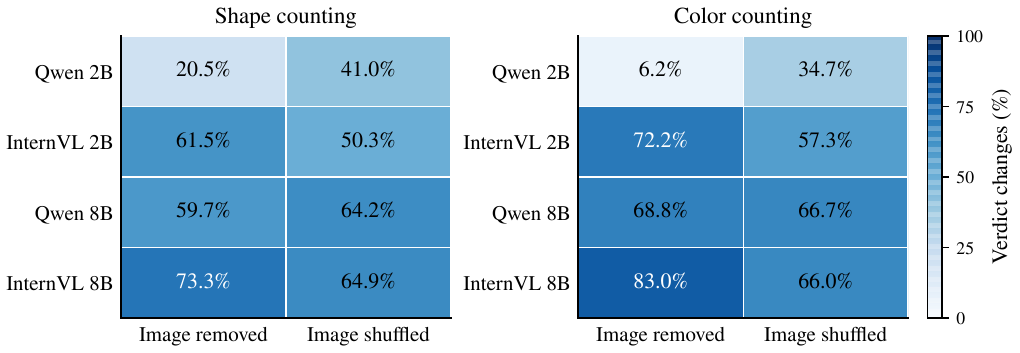}
\caption{Checker verdict changes after image removal or scene shuffling, relative to original images. Each cell uses the same 288 update-selection candidates for its task; values and the shared color scale report percentages. This sensitivity measure does not show whether a changed verdict is correct.}
\label{fig:sensitivity}
\end{figure*}

\begin{figure*}[t]
\centering\includegraphics[width=\textwidth]{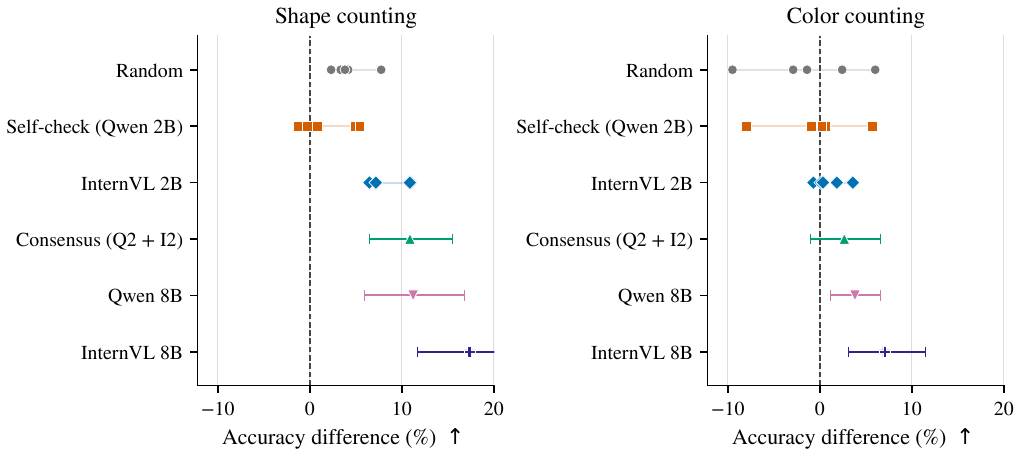}
\caption{Signed held-out-accuracy differences from no update (update $-$ no update, \%). Colored symbols show the five ranking-seed contrasts for the primary rules; original-seed controls show their original-seed contrast with a conditional 95\% scene-group interval. Each fitted update uses 288 unique held-out examples.}
\label{fig:downstream}
\end{figure*}
\section{Results}
\subsection{Checker measurements}
\SensitivityResults{} Figure~\ref{fig:sensitivity} distinguishes image removal from shuffling; these changes do not measure correctness.

\begin{table*}[t]
\centering
\caption{Original-image checker measurements for the update-selection and checker-measurement development splits. Each split has 288 candidates. Precision is correct admissions divided by admissions; false admission is incorrect admissions divided by parseable incorrect candidates; coverage is admissions divided by all candidates. Entries show percentages with numerators and denominators.}
\label{tab:development}
{\nolinenumbers\small\setlength{\tabcolsep}{4pt}
\begin{tabularx}{0.96\textwidth}{@{}llYlll@{}}
\toprule
\textbf{Task} & \textbf{Split} & \textbf{Checker} & \textbf{Precision $\uparrow$} & \textbf{False admit $\downarrow$} & \textbf{Coverage} \\
\midrule
Shape & Update-selection & Qwen 2B & 74.9 (185/247) & 89.9 (62/69) & 85.8 (247/288) \\
Shape & Update-selection & InternVL 2B & 87.6 (155/177) & 31.9 (22/69) & 61.5 (177/288) \\
Shape & Update-selection & Qwen 8B & 100.0 (196/196) & 0.0 (0/69) & 68.1 (196/288) \\
Shape & Update-selection & InternVL 8B & 98.1 (207/211) & 5.8 (4/69) & 73.3 (211/288) \\
Shape & Checker measurement & Qwen 2B & 73.2 (183/250) & 89.3 (67/75) & 86.8 (250/288) \\
Shape & Checker measurement & InternVL 2B & 82.9 (165/199) & 45.3 (34/75) & 69.1 (199/288) \\
Shape & Checker measurement & Qwen 8B & 98.5 (191/194) & 4.0 (3/75) & 67.4 (194/288) \\
Shape & Checker measurement & InternVL 8B & 97.5 (197/202) & 6.7 (5/75) & 70.1 (202/288) \\
Color & Update-selection & Qwen 2B & 79.6 (215/270) & 100.0 (55/55) & 93.8 (270/288) \\
Color & Update-selection & InternVL 2B & 81.7 (170/208) & 69.1 (38/55) & 72.2 (208/288) \\
Color & Update-selection & Qwen 8B & 98.0 (200/204) & 7.3 (4/55) & 70.8 (204/288) \\
Color & Update-selection & InternVL 8B & 90.4 (216/239) & 41.8 (23/55) & 83.0 (239/288) \\
Color & Checker measurement & Qwen 2B & 73.1 (196/268) & 100.0 (72/72) & 93.1 (268/288) \\
Color & Checker measurement & InternVL 2B & 78.8 (182/231) & 68.1 (49/72) & 80.2 (231/288) \\
Color & Checker measurement & Qwen 8B & 92.2 (202/219) & 23.6 (17/72) & 76.0 (219/288) \\
Color & Checker measurement & InternVL 8B & 91.1 (205/225) & 27.8 (20/72) & 78.1 (225/288) \\
\bottomrule
\end{tabularx}}
\end{table*}

\Needspace{10\baselineskip}
Table~\ref{tab:development} compares the four checkpoints on identical original-image candidates across all 576 development inputs per task. \DevelopmentResults{}

\Needspace{24\baselineskip}
\subsection{Complete updates and direct controls}
\begin{table*}[t]
\centering
\caption{Mean held-out accuracy (\%) and signed held-out-accuracy difference (\%) from no update (update $-$ no update). Primary rules show unweighted means across five ranking seeds with descriptive minimum--maximum ranges; checkpoint-specific controls use the original ranking seed and its conditional 95\% scene-group interval. The no-update entry is one shared deterministic reference.}
\label{tab:downstream}
{\nolinenumbers\small\setlength{\tabcolsep}{3.5pt}
\begin{tabularx}{\textwidth}{@{}Yllll@{}}
\toprule
\textbf{Rule} & \textbf{Shape acc. $\uparrow$} & \textbf{Shape difference $\uparrow$} & \textbf{Color acc. $\uparrow$} & \textbf{Color difference $\uparrow$} \\
\midrule
No update & 70.1 & -- & 77.8 & -- \\
Random & 74.4\, [72.5, 77.9] & $+4.3\, [+2.3, +7.8]$ & 76.7\, [68.3, 83.8] & $-1.1\, [-9.5, +6.0]$ \\
Self-check & 72.1\, [68.9, 75.6] & $+1.9\, [-1.3, +5.4]$ & 77.3\, [69.8, 83.4] & $-0.5\, [-8.0, +5.7]$ \\
InternVL 2B & 78.7\, [76.6, 81.1] & $+8.5\, [+6.5, +11.0]$ & 78.8\, [77.1, 81.4] & $+1.0\, [-0.7, +3.6]$ \\
Consensus & 81.0 & $+10.9\, [+6.5, +15.5]$ & 80.4 & $+2.7\, [-1.0, +6.6]$ \\
Qwen 8B & 81.4 & $+11.2\, [+5.9, +16.8]$ & 81.6 & $+3.8\, [+1.2, +6.6]$ \\
InternVL 8B & 87.5 & $+17.4\, [+11.7, +23.3]$ & 84.8 & $+7.1\, [+3.1, +11.5]$ \\
\bottomrule
\end{tabularx}}
\end{table*}

Table~\ref{tab:downstream} reports all seven arms, with five-set summaries for the primary rules and original-seed controls. Figure~\ref{fig:downstream} shows the realized ranking-seed contrasts and the conditional control intervals. \DownstreamResults{}

\Needspace{10\baselineskip}
Improvement over unchanged weights does not by itself establish an advantage over self-checking. \begin{table}[t]
\centering
\caption{Signed held-out-accuracy differences from self-check (update $-$ self-check, \%). Primary entries show the unweighted mean and descriptive minimum--maximum across five ranking seeds; checkpoint-specific controls show the original-seed paired estimate and conditional 95\% scene-group interval. Positive values favor the listed update.}
\label{tab:direct}
{\nolinenumbers\small\setlength{\tabcolsep}{3pt}
\begin{tabularx}{\columnwidth}{@{}Yll@{}}
\toprule
\textbf{Rule} & \textbf{Shape $\uparrow$} & \textbf{Color $\uparrow$} \\
\midrule
Random & $+2.3\, [-1.6, +9.0]$ & $-0.6\, [-3.5, +3.4]$ \\
InternVL 2B & $+6.6\, [+2.2, +12.3]$ & $+1.5\, [-2.1, +8.1]$ \\
Consensus & $+12.2\, [+8.4, +16.1]$ & $+3.6\, [+0.9, +6.4]$ \\
Qwen 8B & $+12.5\, [+7.5, +17.6]$ & $+4.7\, [+2.3, +7.3]$ \\
InternVL 8B & $+18.6\, [+13.4, +24.0]$ & $+8.0\, [+4.5, +11.6]$ \\
\bottomrule
\end{tabularx}}
\end{table}
 Table~\ref{tab:direct} supplies this direct comparison. \DirectComparatorResults{}

\subsection{Consumed supervision and task dependence}
Table~\ref{tab:selection} describes the 48 consumed examples in each realized set, using program answers only for the gate-comparison analysis.

\Needspace{12\baselineskip}
\begin{table}[t]
\centering
\caption{Selected-target precision and correct proposer targets among the 48 consumed examples. Primary entries show the unweighted mean and range over five ranking seeds; controls use the original ranking seed. Precision is agreement between proposer and program answers.}
\label{tab:selection}
{\nolinenumbers\small\setlength{\tabcolsep}{2pt}
\begin{tabularx}{\columnwidth}{@{}llYY@{}}
\toprule
\textbf{Task} & \textbf{Rule} & \textbf{Precision (\%) $\uparrow$} & \textbf{Correct / 48} \\
\midrule
Shape & Random & 77.1\, [72.9, 81.2] & 37.0 [35, 39] \\
Shape & Self-check & 75.4\, [68.8, 83.3] & 36.2 [33, 40] \\
Shape & InternVL 2B & 85.4\, [77.1, 93.8] & 41.0 [37, 45] \\
Shape & Consensus & 81.2\, [81.2, 81.2] & 39.0 [39, 39] \\
Shape & Qwen 8B & 100.0\, [100.0, 100.0] & 48.0 [48, 48] \\
Shape & InternVL 8B & 97.9\, [97.9, 97.9] & 47.0 [47, 47] \\
Color & Random & 81.2\, [75.0, 87.5] & 39.0 [36, 42] \\
Color & Self-check & 80.0\, [72.9, 87.5] & 38.4 [35, 42] \\
Color & InternVL 2B & 81.2\, [77.1, 85.4] & 39.0 [37, 41] \\
Color & Consensus & 83.3\, [83.3, 83.3] & 40.0 [40, 40] \\
Color & Qwen 8B & 100.0\, [100.0, 100.0] & 48.0 [48, 48] \\
Color & InternVL 8B & 93.8\, [93.8, 93.8] & 45.0 [45, 45] \\
\bottomrule
\end{tabularx}}
\end{table}

\SelectionResults{} At the original ranking seed, Qwen 8B selects 48/48 correct proposer targets on both tasks, yet its mean held-out accuracy is lower than InternVL 8B (81.4\% versus 87.5\% on shape; 81.6\% versus 84.8\% on color), whose selected-target precision is 47/48 and 45/48. This descriptive counterexample shows why selected-target precision alone cannot order update utility. These counts describe different training populations; they do not isolate which measured summary caused an accuracy change.

\Needspace{8\baselineskip}
\paragraph{Target-source diagnostic.} We reran the original-seed Random and Self-check updates with the same frozen 48 IDs, images, questions, optimizer settings, and three fitted seeds. Paired proposer-target and program-target runs used the same model checkpoints, training code and software environment, and H100 4g.40gb Multi-Instance GPU (MIG) slice profile; only the training target differed. On shape, the proposer-target updates reached 75.5\% (Random) and 66.8\% (Self-check), while program-target updates reached 78.9\% and 85.2\%; the signed mean held-out-accuracy differences (program target minus proposer target) were +3.5\% and +18.4\%, computed before display rounding. On color, the corresponding values were 80.0\% and 76.6\% versus 84.0\% and 85.2\%; the signed mean held-out-accuracy differences (program target minus proposer target) were +4.1\% and +8.6\%, computed before display rounding. The fitted-seed target-source differences span $-1.4$\% to $+12.5$\% for Random on shape and $+16.3$\% to $+22.6$\% for Self-check; on color they span $+2.8$\% to $+5.2$\% and $+5.6$\% to $+11.5$\%, respectively. Six of the 6,912 predictions in this diagnostic fall outside 0--4 and score as incorrect. This diagnostic isolates target-source sensitivity on these sets; it does not test program-target admission or transfer.

\Needspace{9\baselineskip}
\ReliabilityUtilityResults{} Figure~\ref{fig:relationship} relates these quantities without fitting a trend to six correlated rules.

\Needspace{5\baselineskip}
All rules share one proposer and can overlap in their selected examples. Task-level differences are observations about these fixed experiments, not independent replications of a model-family effect.

Additional fitted-seed accuracies and held-out results by edit type and program answer are reported in Appendix~\ref{app:additional_results}. These descriptive subgroups reuse the same held-out scene groups as the primary analysis.

\Needspace{18\baselineskip}
\FloatBarrier
\section{Discussion}
The experiment distinguishes three populations. Development gate scores concern all candidates at a checker's observed coverage. Selected-target precision concerns the examples retained and actually consumed. Downstream accuracy concerns fresh held-out scenes after a particular optimizer update. A favorable result in one population need not imply a favorable result in another.

\Needspace{12\baselineskip}
Holding the ranking construction, exposure, and optimizer fixed within each realized set strengthens the comparison between admission rules. It does not fix the selected answer distribution or visual composition. The experiment therefore estimates the consequences of each complete rule, including its example selection. It cannot identify supervision correctness as the sole mechanism or establish an architecture or parameter-count effect.

\section{Reproducibility}
The reproduction configuration pins the exact revisions for all four checkpoints.
Supplementary source provides the scene generator, fixed configurations, exact prompts, and recorded predictions. Replay uses recorded outputs; fresh inference requires separate checkpoint access. Qwen uses AutoProcessor; InternVL prepends \texttt{<image>}. Decoding uses one beam, 16 new tokens for development and checking, and 64 for held-out answering. Parsing and interval construction are specified in Sections~3 and~4.

The pinned environment uses PyTorch~\citep{paszke2019pytorch} 2.9.1+cu128, torchvision 0.24.1+cu128, Transformers~\citep{wolf2020transformers} 4.57.1, PEFT 0.18.1, Accelerate 1.14.0, NumPy 2.4.6, and Pillow 11.3.0; supplementary requirements list the remaining direct dependencies. The renderer and recorded-output replay use Python's standard library.

\section{Conclusion}
\ConclusionResults{}

\Needspace{14\baselineskip}
\section*{Limitations}
The study covers two generated counting tasks, one proposer, and four fixed checker checkpoints. The three primary rules use five ranking seeds, while the three original-seed controls use the original ranking seed; every set has three fitted seeds. Group intervals condition on each realized set and its fitted seeds, while the five-seed summaries are descriptive and do not represent a population of selection procedures. Updates are short: one epoch over 48 examples. Selector code omitted program answers and derived correctness fields, but reference answers remained in shared development records accessible to the broader pipeline; we cannot rule out their availability outside the selector. No natural-image or human evaluation was performed.

\Needspace{12\baselineskip}
\section*{Ethical Considerations}
Potential downstream risks arise if answer-admission gates curate training data beyond these synthetic tasks. Incorrect admissions or rejected valid examples could propagate errors or alter later training populations. Our results do not establish safety, fairness, or utility on natural images or in human-facing deployments. The generated scenes, counting prompts, integer answers, and YES/NO verdicts contain no personal information, third-party photographs, or objectionable human-derived content. No participants, annotators, or manual ratings were used. Participant instructions, recruitment, payment, participant consent, and annotator demographics are therefore inapplicable; we make no claim of ethics-board approval or exemption.

\paragraph{Checkpoint and reuse rights.} The repositories for the four fixed checkpoints specify Apache-2.0 terms. The reproduction package provides authored source, configurations, generated labels, and recorded counting outputs; it distributes no model weights, third-party image data, or dependency source or binaries. Live model reproduction obtains checkpoints and software separately under their upstream terms. No separate reuse license is supplied for the authors' code or generated data.

\paragraph{Computational resources.} Recorded allocations for the reported experiments were 25.19 A100-SXM4-40GB accelerator-hours for the selection-seed matrix, 2.93 H100 4g.40gb MIG slice-hours for the target-source diagnostic, and 5.27 full-device H100 hours for complete-epoch development. These resource types are reported separately and are not a full-device-hour equivalent. They do not cover every exploratory or setup activity, so they are not an exhaustive project total. Allocation is not utilization; energy and emissions were not measured.

\section*{Acknowledgments} OpenAI Codex assisted with literature discovery, experiment planning, code development, analysis, and manuscript revision. The authors reviewed the sources, code, analyses, and prose and take responsibility for the final paper.
\bibliography{references}
\clearpage
\suppressfloats[t]
\appendix
\section{Additional per-seed and subgroup results}\label{app:additional_results}
Table~\ref{tab:seeds} reports the fitted-seed accuracies underlying the update summaries. Tables~\ref{tab:edits} and~\ref{tab:answers} report descriptive held-out accuracies by edit type and program answer. The groups reuse held-out scenes and do not constitute independent replications; the subgroup results carry no separate population-level intervals.
Table~\ref{tab:seeds} shows substantial fit-seed variation among the primary rules, with shape accuracy spanning 58.3--84.4\% and color accuracy spanning 59.0--87.5\% across their 15 fitted updates. Table~\ref{tab:edits} shows that InternVL 8B has the highest displayed accuracy in all six task-by-edit subgroups (86.5--88.2\% on shape and 83.7--86.1\% on color). Table~\ref{tab:answers} shows near-ceiling accuracy for answer 0 and lower accuracy for answer 4 (19.4--63.0\% on shape and 24.1--60.2\% on color). Each task has 36 unique examples for answers 0 and 4 and 72 for answers 1--3; these descriptive groups reuse held-out scenes.
\begin{table}[t]
\centering
\caption{Accuracy (\%) over fitted training seeds. Primary entries give the minimum--maximum across the 15 fitted updates (five ranking seeds and three fitted seeds); controls give the three fitted-seed values at the original ranking seed.}
\label{tab:seeds}
{\nolinenumbers\small\setlength{\tabcolsep}{3pt}
\begin{tabularx}{\columnwidth}{@{}Yll@{}}
\toprule
\textbf{Rule} & \textbf{Shape $\uparrow$} & \textbf{Color $\uparrow$} \\
\midrule
Random & 67.4--84.4 & 63.9--86.8 \\
Self-check & 58.3--83.7 & 59.0--87.5 \\
InternVL 2B & 70.8--84.4 & 71.9--83.7 \\
Consensus & 82.3 / 75.0 / 85.8 & 80.2 / 80.6 / 80.6 \\
Qwen 8B & 73.6 / 86.8 / 83.7 & 81.9 / 85.1 / 77.8 \\
InternVL 8B & 90.6 / 87.8 / 84.0 & 86.1 / 82.3 / 86.1 \\
\bottomrule
\end{tabularx}}
\end{table}

\begin{table}[t]
\centering
\caption{Held-out accuracy (\%) by edit type: clean/answer-changing/answer-preserving. Primary entries average the same 15 fitted updates used in the five-seed summary; controls use the original ranking seed. Slash-separated values are in the edit-type order shown above: clean/answer-changing/answer-preserving; checkpoint-specific controls average their three fitted seeds at the original ranking seed. Subgroups are descriptive.}
\label{tab:edits}
{\nolinenumbers\small\setlength{\tabcolsep}{3pt}
\begin{tabularx}{\columnwidth}{@{}Yll@{}}
\toprule
\textbf{Rule} & \textbf{Shape $\uparrow$} & \textbf{Color $\uparrow$} \\
\midrule
No update & 68.8 / 71.9 / 69.8 & 77.1 / 79.2 / 77.1 \\
Random & 73.9 / 74.4 / 75.0 & 76.9 / 76.9 / 76.3 \\
Self-check & 71.7 / 71.7 / 72.8 & 76.9 / 77.8 / 77.2 \\
InternVL 2B & 77.8 / 79.2 / 79.0 & 78.8 / 79.6 / 78.1 \\
Consensus & 80.2 / 80.9 / 81.9 & 80.2 / 79.9 / 81.2 \\
Qwen 8B & 81.2 / 81.2 / 81.6 & 79.9 / 83.0 / 81.9 \\
InternVL 8B & 86.5 / 88.2 / 87.8 & 83.7 / 84.7 / 86.1 \\
\bottomrule
\end{tabularx}}
\end{table}

\begin{table}[t]
\centering
\caption{Held-out accuracy (\%) by program answer. Primary values average 15 fitted updates; controls average three. For answers 0 through 4, support is 36/72/72/72/36 unique examples per task (24/48/48/48/24 scene groups). Counts refer to unique examples, not repeated predictions.}
\label{tab:answers}
{\nolinenumbers\small\setlength{\tabcolsep}{2pt}
\begin{tabularx}{\columnwidth}{@{}Yrrrrr@{}}
\toprule
\textbf{Rule} & \textbf{0 $\uparrow$} & \textbf{1 $\uparrow$} & \textbf{2 $\uparrow$} & \textbf{3 $\uparrow$} & \textbf{4 $\uparrow$} \\
\midrule
\multicolumn{6}{@{}l@{}}{\textbf{Shape}} \\
\midrule
No update & 100.0 & 94.4 & 70.8 & 55.6 & 19.4 \\
Random & 100.0 & 96.5 & 83.2 & 57.3 & 21.3 \\
Self-check & 100.0 & 95.3 & 78.7 & 54.1 & 20.6 \\
InternVL 2B & 100.0 & 96.8 & 84.9 & 66.9 & 32.2 \\
Consensus & 100.0 & 97.7 & 89.4 & 69.9 & 34.3 \\
Qwen 8B & 100.0 & 93.5 & 90.3 & 63.9 & 55.6 \\
InternVL 8B & 100.0 & 98.1 & 91.7 & 78.7 & 63.0 \\
\multicolumn{6}{@{}l@{}}{\textbf{Color}} \\
\midrule
No update & 100.0 & 100.0 & 76.4 & 63.9 & 41.7 \\
Random & 100.0 & 100.0 & 76.4 & 63.9 & 33.1 \\
Self-check & 100.0 & 100.0 & 75.7 & 66.9 & 33.1 \\
InternVL 2B & 99.8 & 100.0 & 77.3 & 72.4 & 31.3 \\
Consensus & 100.0 & 100.0 & 77.3 & 82.4 & 24.1 \\
Qwen 8B & 100.0 & 100.0 & 81.5 & 69.0 & 51.9 \\
InternVL 8B & 100.0 & 100.0 & 81.0 & 78.2 & 60.2 \\
\bottomrule
\end{tabularx}}
\end{table}

\end{document}